\pdfoutput=1

\documentclass[11pt]{article}

\usepackage[]{acl}

\usepackage{times}
\usepackage{latexsym}

\usepackage[T1]{fontenc}

\usepackage[utf8]{inputenc}

\usepackage{microtype}

\usepackage{graphicx}
\usepackage{ctable}
\usepackage{amsfonts}
\usepackage{multirow}
\usepackage{hyperref}
\usepackage{amsmath}
\usepackage{xcolor}
\usepackage{transparent}
\usepackage{stmaryrd}

\title{Emp-RFT: Empathetic Response Generation via Recognizing Feature Transitions between Utterances}

\author{Wongyu Kim$^{1}$, Youbin Ahn$^{2}$, Donghyun Kim$^{2}$, and Kyong-Ho Lee$^{2}$\\
	$^{1}$Department of Artificial Intelligence, $^{2}$Department of Computer Science \\
	Yonsei University, Seoul, Republic of Korea \\
	\texttt{\{rladnjsrb9999, ybahn, dhkim92, khlee98\}@yonsei.ac.kr} \\
	}

\begin{document}
	\maketitle
	\begin{abstract}
		Each utterance in multi-turn empathetic dialogues has features such as emotion, keywords, and utterance-level meaning. Feature transitions between utterances occur naturally. However, existing approaches fail to perceive the transitions because they extract features for the context at the coarse-grained level. To solve the above issue, we propose a novel approach of recognizing feature transitions between utterances, which helps understand the dialogue flow and better grasp the features of utterance that needs attention. Also, we introduce a response generation strategy to help focus on emotion and keywords related to appropriate features when generating responses. Experimental results show that our approach outperforms baselines and especially, achieves significant improvements on multi-turn dialogues.
	\end{abstract}
	
	\section{Introduction}
	\label{sec:intro}
\begin{figure}[t]
		\centering
		\includegraphics[width=\columnwidth]{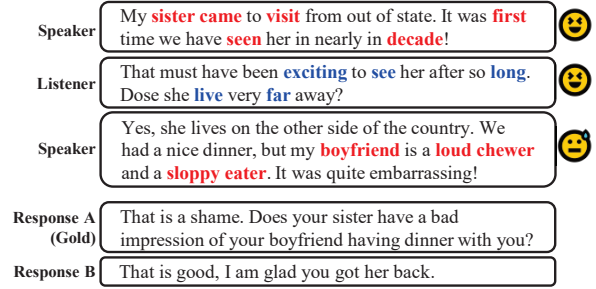}
		\caption{An example of EmpatheticDialogues with response A and B. Response B is from one of state-of-the-art models. Highlighted words are keywords.}
		\label{fig:motivating}
	\end{figure}

	Humans have empathy which is the ability to understand situations others have experienced and emotions they have felt from the situations \cite{eisenberg1987critical}. That ability also enables to interest and console others while sharing a conversation. Thus, empathetic response generation task has been considered noteworthy. Figure \ref{fig:motivating} shows an example of a multi-turn empathetic dialogue dataset, EmpatheticDialogues \cite{rashkin2019towards} constructed to solve the task. A speaker talks about one of 32 emotion labels and a situation related to the emotion label, and a listener empathizes, responding to the speaker. Existing approaches \cite{rashkin2019towards, lin2019moel, majumder2020mime, li2020empdg, kim2021perspective} for the task achieve promising results but show limitations when dialogues become long because they extract features from the concatenation of all tokens in the context at the coarse-grained level.
	
	However, at the fine-grained level, each utterance in multi-turn empathetic dialogues has features such as emotion, keywords that each denote what an interlocutor feels and primarily says, and utterance-level meaning that can be known when looking at the entire utterance. In addition, it is a natural phenomenon that features of each utterance differ from the previous, as the dialogue is prolonged. Hence, we humans instinctively recognize these feature transitions, which helps us understand how the dialogue flows and grasp the features of utterance that needs attention. Also, humans respond to others, focusing on emotion and keywords related to appropriate features. Take the example in Figure \ref{fig:motivating}. In the first turn, the speaker is excited to see the speaker's sister in a long time by mentioning keywords (e.g., ‘sister’, ‘visit’, ‘decade’), and the listener reacts to the excitement and asks about her by mentioning keywords (e.g., ‘exciting’, ‘see’, ‘live’). However, in the second speaker utterance, the speaker becomes embarrassed because of the speaker's boyfriend’s bad table manners by mentioning keywords (e.g., ‘boyfriend’, ‘loud’, ‘eater’). We humans recognize that the features of second speaker utterance have changed compared to those of previous utterances, and usually decide to be attentive to the features of the second utterance. Then, by focusing on information such as keywords of that utterance and emotion and keywords (e.g., ‘bad', ‘impression') related to the features of that utterance, humans generate empathetic, coherent, and non-generic responses like response A. However, the model which produces non-empathetic and incoherent response like response B, considers that the features of the first speaker utterance represent the context from the coarse-grained view. 
	
	In this paper, we first propose to annotate features on each utterance at the fine-grained level (§\ref{sec:datapre}). Then, we introduce a novel \textbf{Emp}athetic response generator based on \textbf{R}ecognizing \textbf{F}eature \textbf{T}ransitions (Emp-RFT), which has two essential parts: Feature Transition Recognizer and Response Generation Strategy. The first part recognizes feature transitions between utterances, utilizing comparison functions of \citet{wang2017compare}, which makes Emp-RFT understand the dialogue flow and grasp appropriate features of utterance that needs attention. The second part helps Emp-RFT focus on emotion and keywords related to appropriate features. Specifically, by fusing context with keywords, such keywords are emphasized within each utterance and get more attention when generating responses. Then, Emp-RFT detects next emotion and keywords that denote emotion and keywords of the response, which helps figure out proper emotion and keywords for generation. Lastly, inspired by \citet{Dathathri2020Plug,chen2020simple}, a new mechanism of Plug and Play Language Model(PPLM), contrastive PPLM using contrastive loss, is introduced, which controls Emp-RFT to actively use the keywords detected to be next keywords when generating responses.
	
	We conduct experiments on EmpatheticDialogues. Emp-RFT outperforms strong baselines, particularly, when dialogues are multi-turn.
	
	Our main contributions are as follows. (1) We introduce a novel approach that recognizes feature transitions between utterances, which results in understanding how the dialogue flows and grasping the features of utterance that the model should be attentive to. (2) We propose a response generation strategy including fusing context with keywords, next emotion and keywords detection, and contrastive PPLM. The strategy makes our model focus on emotion and keywords related to appropriate features when generating responses. (3) In the experiments, Emp-RFT outperforms baselines, especially, when dialogues are prolonged.
	
	\section{Related Work}
	\label{sec:related}
	
	Since \citet{rashkin2019towards} release EmpatheticDialogues, many approaches have been proposed to generate empathetic responses. \citet{lin2019moel} propose mixture of emotional experts. \citet{majumder2020mime} propose emotion grouping, emotion mimicry, and stochastic sampling. \citet{li2020empdg} extract emotional words through lexicon and propose an adversarial generative model. \citet{shen2021constructing} apply dual-learning with unpaired data for the bidirectional empathy. \citet{gao2021improving} integrate emotion cause into response generation process through gated mechanism. \citet{sabour2021cem,li-etal-2022-kemp} use implicit commonsense for context modelling. \citet{kim2021perspective} train a model to extract words that cause the speaker's emotion and attach RSA Framework \cite{frank2012predicting} to any generative models to generate responses, focusing on emotion cause words.
	
	Recently, many studies have shown remarkable improvements through recognizing transitions of features between utterances in open-domain multi-turn dialogues. \citet{qiu2020if} perceive transitions of emotion states for context modelling. \citet{zou2021thinking} propose a module to manage keyword transitions. \citet{zhan2021augmenting} model external knowledge transitions to select a knowledge used for generation. In multi-turn empathetic dialogues, we consider emotions, keywords, and utterance-level meaning \cite{gu2021dialogbert} as important features of each utterance and propose a novel approach of recognizing feature transitions between utterances.

	\section{Task Formulation}
	\label{sec:task}
	Given context $con=[u^{1},…,u^{n-1}]$, where an utterance $u^{i}=[u^{i}_{1},…,u^{i}_{|u^{i}|}]$ consists of $|u^{i}|$ words, we can obtain $e=[e^{1},…,e^{n-1}]$ and $k=[k^{1},...,k^{n-1}]$, where $e^{i}$ and $k^{i}=[k^{i}_{1},...,k^{i}_{|k^{i}|}]$ each denote emotion and $|k^{i}|$ keywords of $u^{i}$ through data preparation (§\ref{sec:datapre}). To conduct next keywords detection, we construct keyword pairs $kps$ (§\ref{sec:keypairs}) whose each pair has two keywords each from keywords of the speaker utterance and keywords of the listener utterance in the same turn. Finally, given $con, e, k$, and $kps$, we detect next emotion $e^{y}$ and next keywords $k^{y}=[k^{y}_{1},…,k^{y}_{|k^{y}|}]$, and generate an empathetic response $y=[y_{1},…,y_{m}]$.
	
	\begin{table}[t]
		\centering
		\resizebox{\columnwidth}{!}{%
			{\small
				\begin{tabular}{cccc}
					\specialrule{.08em}{.05em}{.2em}
					\textbf{Feature} & \textbf{Top-1 Acc} & \textbf{Top-5 Acc} & \textbf{macro-F1}  \\ 
					\specialrule{.08em}{.05em}{.2em}
					\textbf{EofSU}           & 46.77            & 81.26            & 43.55     \\
					\specialrule{.08em}{.05em}{.2em}
					\textbf{EofLU}            & 58.44           & 89.96           & 53.25           \\
					\specialrule{.08em}{.05em}{.2em}
					\specialrule{.08em}{.05em}{.2em}
					\textbf{Feature} & \textbf{TL-P} & \textbf{TL-R} & \textbf{TL-F1}  \\ 
					\specialrule{.08em}{.05em}{.2em}
					\textbf{KofSU}         & 41.53            & 66.58           & 51.15    \\
					\specialrule{.08em }{.05em}{.2em}
					\textbf{KofLU}           & 52.31           & 60.97           & 56.30            \\
					\specialrule{.08em}{.05em}{.2em}
				\end{tabular}%
			}
		}
		\caption{Performances of feature annotations. When we evaluate annotations of \textbf{EofSU}/\textbf{KofSU}, EMOCAUSE \cite{kim2021perspective} made based on EmpatheticDialogues, is used to verify emotion and emotion cause words detection models. For evaluations of keyword annotations, we use the metrics, Token-Level(\textbf{TL}) Precision(\textbf{P}), Recall(\textbf{R}), and \textbf{F1} \cite{deyoung2020eraser}, usually used in token extraction tasks.}
		\label{table:anno1}
	\end{table}

	\section{Data Preparataion}
	\label{sec:datapre}
	In this section, we introduce feature annotation in the speaker and listener utterances. 
	\subsection{Feature Annotation in Speaker Utterances}
	\label{sec:featanno}
	
	\textbf{Emotion and Keywords of Speaker Utterance} \textbf{(EofSU/KofSU).} Speakers try to say an emotional experience that causes a certain emotion in the utterance. Thus, we leverage a model \cite{kim2021perspective} which is trained to jointly detect an emotion and emotion cause words of the speaker utterance, using EmpatheticDialogues. We regard top-6 emotion cause words as keywords and remove stopwords and punctuations in keywords.

	\subsection{Feature Annotation in Listener Utterances}
	\label{sec:keypairs}
	
	\textbf{Emotion of Listener Utterance} \textbf{(EofLU).} We finetune RoBERTa \cite{liu2019roberta} to detect an emotion given a situation description in EmpatheticDialogues. Then, the model predicts an emotion of the listener utterance.
	
	\textbf{Keywords of Listener Utterance} \textbf{(KofLU).} Listeners express empathy in the utterance through three Communication Mechanisms (CMs) \cite{sharma2020computational} including emotional reaction, interpretation, and exploration. Thus, three models are leveraged, where each model is trained to detect words that cause one of three CMs, using another dialogue dataset for mental health support \footnote{A dialogue has a (post, response) pair, and words which cause each CM are annotated on each dialogue.}. Then, three models predict such words in the listener utterance. Since predicted words take up slightly a lot in the listener utterance, these words are filtered out in the keyword pairs construction.
	
	\textbf{Keyword Pairs Construction.} Inspired by \citet{zou2021thinking}, keyword pairs $kps$ are constructed not only to filter out above predicted words, but also to conduct next keyword detection. Given a dialogue corpus, all pairs are extracted, where each pair has a head word and a tail word each from keywords in the speaker utterance and predicted words in the listener utterance in the same turn. Then, all pairs are filtered out to obtain high-frequency pairs through pointwise mutual information (PMI)\footnote{We use pairs whose PMI $\ge$ 1. The pairs whose tail words are stopwords or punctuations, are removed.} \cite{church1990word} which can measure the association between two words in a corpus. Filtered pairs become $kps$. A tail word of a $kp$ is regarded as a keyword of the listener utterances joined to extract that keyword pair.
	
	Performances of feature annotations are summarized in Table \ref{table:anno1} and show reliable results. However, test sets for \textbf{KofLU} based on EmpatheticDialogues, don't exist. Thus, we randomly sample 100 test dialogues in EmpatheticDialogues and ask 3 human workers to annotate whether each word plays important role for empathizing in the listener utterances. By majority voting, the final verdict on each annotation is decided.
	We compute the inter-annotator agreement on annotation of test sets for \textbf{KofLU} through Fleiss' kappa ($\boldsymbol{\kappa}$) \cite{fleiss1973equivalence}, and result in 0.55, where $0.4 < \boldsymbol{\kappa} < 0.6$ indicates moderate agreement.
	
	\begin{figure}[t]
		\centering
		\includegraphics[width=\columnwidth]{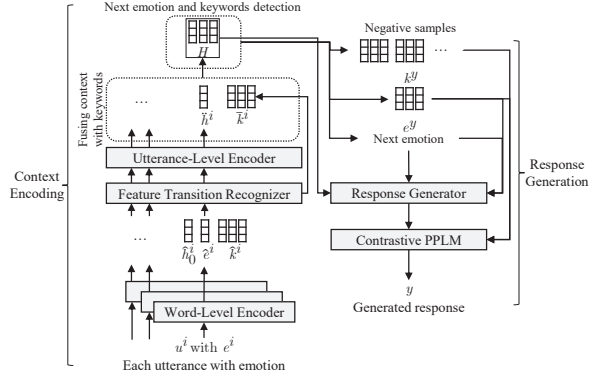}
		\caption{Overall architecture of Emp-RFT.}
		\label{fig:arch}
	\end{figure}

	\section{The Emp-RFT Model}
	\label{method}
	In this section, we detail Emp-RFT whose overall architecture is shown in Figure \ref{fig:arch}.
	
	\subsection{Context Encoding}
	\label{method:context}
	\textbf{Word-Level Encoder.} Emp-RFT contains an encoder $f_{\theta}(\cdot)$ which has the six-layer encoder of BART \cite{lewis2020bart} as the backbone and extracts feature vectors of each $u^{i}$. Inspired by BERT \cite{devlin2019bert}, we prefix each utterance with a $[SEN]$ token, so $u^{i}_{0} = [SEN]$. Then, each token is represented as $emb^{i}_{j}$, the sum of the following four embeddings: word embedding, position embedding, role embedding and emotion embedding $M_{e}\in\mathbb{R}^{n_{emo} \times d}$ \footnote{$j$, $d$, and $n_{emo}$ each denote the index for words, hidden size, and the number of emotion classes. The role and emotion embbedings are each for distinguishing two interlocutors and for incorporating the emotion into each utterance.}. Then, the encoder transforms each utterance into a list of output hidden states:
	\begin{equation}
		[\hat{h}^{i}_{0},...,\hat{h}^{i}_{|u^{i}|}]=f_{\theta}([emb^{i}_{0},...,emb^{i}_{|u^{i}|}]),
	\end{equation}
	where $\hat{h}^{i}_{j}\in\mathbb{R}^{d}$. For each utterance, we can obtain utterance-level meaning vector $\hat{h}^{i}_{0}$ derived from the token $[SEN]$, concatenated keyword vectors $\hat{k}^{i}\in\mathbb{R}^{|k^{i}|\times d}$ derived from the tokens corresponding to $k^{i}_{p}$ ($p$ is the index for keywords.), and emotion vector $\hat{e}^{i}=M_{e}\hat{h}^{i}_{0}$.

\begin{figure}[t]
		\centering
		\includegraphics[width=\columnwidth]{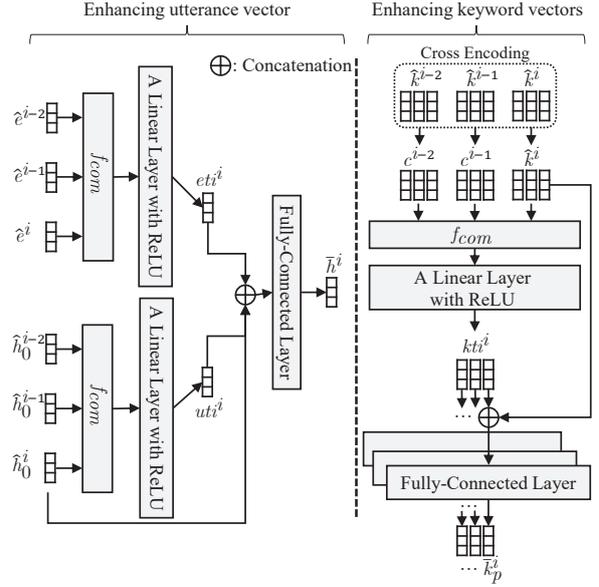}
		\caption{Operation process of feature transition recognizer.}
		\label{fig:ftr}
	\end{figure}

	\textbf{Feature Transition Recognizer.} Emp-RFT has a component that operates as the process illustrated in Figure \ref{fig:ftr}. The component computes feature transition information between feature vectors, utilizing two comparison functions, subtraction and multiplication of \citet{wang2017compare}. Each feature vector is compared to previous two feature vectors \footnote{If there aren't previous feature vectors, we can obtain those by regarding first output hidden state of a padded utterance as utterance and keyword vectors \cite{qiu2020if}.}. First, emotion transition information $eti^{i}$ is computed:
	\begin{equation}
		\label{eq45}
		eti^{i}=\mathrm{ReLU}(W_{eti}(f_{com}(\hat{e}^{i},\hat{e}^{i-1},\hat{e}^{i-2}))),
	\end{equation}
	\begin{equation}
		\begin{aligned}
			&f_{com}(\hat{e}^{i},\hat{e}^{i-1},\hat{e}^{i-2}) \\
			&=\begin{bmatrix}
				(\hat{e}^{i}-\hat{e}^{i-1})\odot(\hat{e}^{i}-\hat{e}^{i-1})\\
				\hat{e}^{i}\odot \hat{e}^{i-1}\\
				(\hat{e}^{i}-\hat{e}^{i-2})\odot(\hat{e}^{i}-\hat{e}^{i-2})\\
				\hat{e}^{i}\odot \hat{e}^{i-2}
			\end{bmatrix},
		\end{aligned}
	\end{equation}
	where $f_{com}$ and $\odot$ each denote our transition information computing function and Hadamar product, and $W_{eti}\in\mathbb{R}^{d \times {4n_{emo}}}$. Next, utterance-level meaning transition information $uti^{i}$ is computed:
	\begin{equation}
		\label{eq67}
		uti^{i}=\mathrm{ReLU}(W_{uti}(f_{com}(\hat{h}^{i}_{0},\hat{h}^{i-1}_{0},\hat{h}^{i-2}_{0}))),
	\end{equation}
	where $W_{uti}\in\mathbb{R}^{d \times {4d}}$. We then obtain enhanced utterance vector of each utterance by integrating utterance-level meaning vector, and emotion and utterance-level meaning transition information: 
	\begin{equation}
		\bar{h}^{i}=\mathrm{FC}_{utt}([\hat{h}^{i}_{0};eti^{i};uti^{i}]),
	\end{equation}
	where $FC_{utt}$ is a fully-connected layer with size of $d$. In addition, keyword transition information $kti^{i}$ is computed between concatenated keyword vectors and cross-encoded vectors $c^{t}$, where $t\in\{i-1,i-2\}$:
	\begin{flalign}
		\label{eq9101112}
		&kti^{i}=\mathrm{ReLU}(W_{kti}(f_{com}(\hat{k}^{i},c^{i-1},c^{i-2}))^{T}), \\
		&c^{t}=\mathrm{softmax}(Q^{i}(K^{t})^{T})\hat{k}^{t},\\
		&Q^{i} = \hat{k}^{i}W_{Q}, K^{t} = \hat{k}^{t}W_{K},
	\end{flalign}
	where $W_{kti} \in \mathbb{R}^{d\times 4d}, W_{Q}$ and $W_{K} \in \mathbb{R}^{d\times d}$. We can obtain enhanced keyword vector of each keyword by integrating keyword vector, and keyword transition information: 
	\begin{equation}
		\bar{k}^{i}_{p}=\mathrm{FC}_{key}([\hat{k}^{i}_{p};kti^{i}_{p}]),
	\end{equation}
	where $FC_{key}$ is a fully-connected layer with size of $d$. Consequently, the enhanced feature vectors guide Emp-RFT to accurately grasp the features of utterance that the model should be attentive to when given feature transition information.
	
	\textbf{Utterance-Level Encoder.} Emp-RFT contains another encoder $g_{\phi}(\cdot)$ which has the six-layer encoder of BART, and transforms enhanced utterance vectors with global position embeddings (GPE) into a context representation to capture relationships between utterances \cite{gu2021dialogbert}:
	\begin{equation}
		[\ddot{h}^{1},...,\ddot{h}^{n-1}]=g_{\phi}([\bar{h}^{1},...,\bar{h}^{n-1}]).
	\end{equation}
	Emp-RFT consists of hierarchical structures of encoders through word-level and utterance-level encoders. This structure makes Emp-RFT comprehend each utterance at the fine-grained level, and understand the context by integrating information based on comprehension of each utterance.
	
\begin{figure}[t]
		\centering
		\includegraphics[width=\columnwidth]{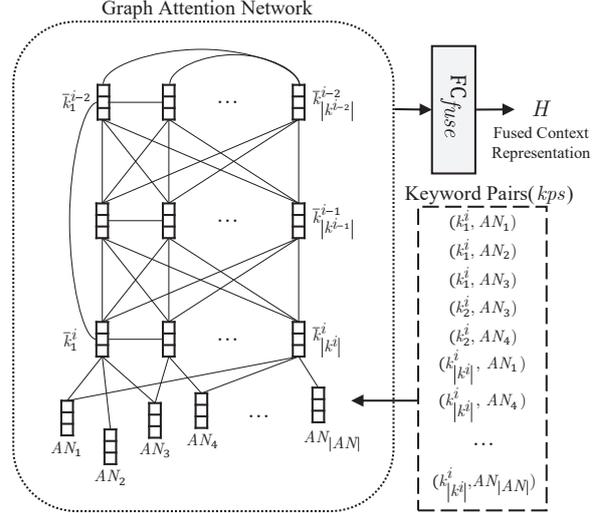}
		\caption{Operation process of fusing context with keywords. $AN_{o}$ is the appended node. Some symbols and edges are omitted for simplicity.}
		\label{fig:fck}
	\end{figure}

	\textbf{Fusing Context with Keywords.} Emp-RFT fuses context with keywords as the process illustrated in Figure \ref{fig:fck}. We first dynamically build keyword graph for each context. Keywords in each context become nodes and are initialized by corresponding enhanced keyword vectors with GPE. Edges are built across the below cases: (1) between two keywords from the same utterance and (2) between a keyword from a certain utterance and another keyword from the previous two utterances. Also, a tail word in a $kp$ whose head word is $k^{n-1}_{p}$ is appended as a node and connected with $k^{n-1}_{p}$ node. Appended nodes ($ANs$) are initialized through BART decoder whose parameters are frozen with GPE, and used for next keywords detection. To obtain keyword representation $\hat{v}^{i}_{o}$ from the keyword graph($o$ is the index for nodes.), nodes are updated based on multi-head graph-attention mechanism  \cite{velickovic2018graph, li-etal-2022-kemp}. This mechanism makes Emp-RFT not only capture relationships between nodes but also manage influences of each appended node through attention architecture:
	\begin{flalign}\label{eq1516}
		&\hat{v}^{i}_{o}=v^{i}_{o}+ \bigparallel_{mh=1}^{MH}\sum_{z\in A^{i}_{o}}\alpha^{i,mh}_{oz}(W^{mh}_{v}v_{z}), \\
		&\alpha^{i,mh}_{oz}={\mathrm{exp}((W^{mh}_{q}v^{i}_{o})^{T}W^{mh}_{key}v_{z})\over\sum_{s\in\mathcal{A}^{i}_{o}}\mathrm{exp}((W^{mh}_{q}v^{i}_{o})^{T}W^{mh}_{key}v_{s})},
	\end{flalign}
	where $v^{i}_{o}$, $\Arrowvert$, $A^{i}_{o}$, and $\alpha^{i,mh}_{oz}$ each denote a node representation, the concatenation of $MH$ attention heads, the neighbours of $v^{i}_{o}$ in the adjacency matrix $A$, and self-attention weight and $W^{mh}_{v}$, $W^{mh}_{q}$, $W^{mh}_{key} \in \mathbb{R}^{d_{mh}\times d}$ ($d_{mh}=d/MH$). Lastly, we can obtain the fused context representation $H=[h^{1},...,h^{n-1}]$ by fusing the context representation with the sum of keyword representations:
	\begin{equation}
		{h}^{i}=\mathrm{FC}_{fuse}([\ddot{h}^{i};\mathrm{sum}([\hat{v}^{i}_{1},...,\hat{v}^{i}_{|k^{i}|}])]),
	\end{equation} where $FC_{fuse}$ is a fully-connected layer with size of $d$.
	Consequently, keywords are emphasized within each utterance and get greater attention when generating responses.
	
	\textbf{Next Emotion and Keywords Detection.} Emp-RFT detects next emotion $e^{y}$ and keywords $k^{y}$, which helps figure out proper emotion and keywords for generation. First, based on the maxpooled fused context representation, next emotion distribution is predicted: 
	\begin{equation}
		\label{a1}
		P_{e}=\mathrm{softmax}(M_{e}\mathrm{MP}(H)),
	\end{equation}
	where $\mathrm{MP}$ denotes maxpooling. We use the emotion with the highest probability ($\hat{e}^{y}$) for generation. Also, Emp-RFT predicts whether the word of each $AN$ belongs to the next keywords through the binary classification, where the true label denotes the word belongs to: 
	\begin{equation}
		\label{a2}
		P_{k}=\prod_{o=1}^{|ANs|}\mathrm{softmax}(W_{AN}[\hat{v}^{n}_{o};\mathrm{MP}(H)]),
	\end{equation}
	where $W_{AN}\in \mathbb{R}^{2\times 2d}$. We consider the words of $ANs$ whose probabilities for the true label $\ge$ $0.8$ as the keywords ($\hat{k}^{y}$) for generation.
	
	\subsection{Response Generation}
	\label{method:response}
	\textbf{Response Generator.} Emp-RFT includes a response generator (RG) which has the six-layer decoder of BART as the backbone. Through the four embeddings with $\hat{e}^{y}$, explained previously, we can obtain the input sequence embedding for RG. We prefix it with the sum of node representations corresponding to $\hat{k}^{y}$. Then, RG is fed to predict probability distribution on each next token $y_{t}$ based on the fused context representation:
	\begin{equation}
		\label{a3}
		P(y|con,e,k,kps)=\prod_{t=1}^{m}P(y_{t}|y_{<t}, H).
	\end{equation}
	
	\textbf{Training.} We apply cross-entropy loss to three objectives (eq. \ref{a1}, \ref{a2}, \ref{a3}), and train parameters of Emp-RFT in end-to-end manner through the sum of all losses .
	
	\textbf{Contrastive PPLM.} Analysis on the generated responses of the trained Emp-RFT shows that an active reflection of $\hat{k}^{y}$ is demanded. Thus, inspired by \citet{Dathathri2020Plug, chen2020simple}, we propose Contrastive PPLM with a discriminator using contrastive loss. Existing discriminators \cite{Dathathri2020Plug,majumder2021ask} are trained to predict whether a sentence contains a certain attribute, using cross-entropy loss. Then, the gradient of the loss is passed to the generative model to generate a sentence containing such attribute during inference. However, since keywords are not attributes but objects, we train a discriminator to predict whether a response in EmpatheticDialogues is more similar to the keyword set of the response(positive sample) than the keyword sets of another responses(negative samples) in the same batch, using contrastive loss based on the similarity between objects:
	\begin{equation}\label{eq26}
		L^{a}_{pplm}=-\mathrm{log}{\mathrm{exp}(r^{T}_{a}ks_{a}/\tau) \over \sum_{b=1}^{B}\mathrm{exp}(r^{T}_{a}ks_{b}/\tau)},
	\end{equation}
	where $r$,$ks$,$\tau$ and $B$ each denote response and keyword set representations, a temperature parameter and batchsize. During inference, we repeatedly sample three random $ANs$ except for nodes of $\hat{k}^{y}$, and consider the sum of such $AN$ representations as one of negative samples and the sum of node representations corresponding to $\hat{k}^{y}$ as a positive sample. Then, the gradient of the contrastive loss is passed to Emp-RFT.

	\section{Experiments}
	\label{method:experiment}
	
	\subsection{Dataset and Baselines}
	\textbf{Dataset.} Experiments were conducted on EmpatheticDialogues \cite{rashkin2019towards} which contains 24,850 multi-turn dialogues. For each dialogue, we can extract a certain number of instances corresponding to the number of turns within the dialogue. This totals to 47,611 instances, where 22,761 are multi-turn. In one turn of a dialogue, a speaker talks about one of 32 evenly distributed emotion labels and a situation related to the emotion label and a listener empathizes by responding to the speaker. Following the instructions of the dataset, we use 8:1:1 train/valid/test split.
	
	\textbf{Baselines.} We compared Emp-RFT to the following five baseline models: (1) \textbf{MoEL} \cite{lin2019moel} is a transformer-based generative model, which has decoders for each emotion and integrates outputs of the decoders according to predicted emotion distribution. (2) \textbf{EmpDG} \cite{li2020empdg} uses emotional words and consists of an adversarial framework including a generator and discriminators which reflect the user feedback. (3) \textbf{MIME} \cite{majumder2020mime} is also a transformer-based generative model which mimics user emotion based on emotion grouping and uses stochastic sampling for varied responses. (4) \textbf{MIME+Focused S1} and (5) \textbf{Blender+Focused S1} \cite{kim2021perspective} attach RSA Framework to MIME and Blender \cite{roller2021recipes}. Blender is a pretrained model with 90M parameters size, using an immense number of dialogues. It is finetuned on EmpatheticDialogues. Using distractors and Bayes' Rules, RSA Framework makes the models focus on certain parts of the post, such as emotion cause words when generating responses in the single-turn dialogues \footnote{To make the models work in the multi-turn dialogues, the models are converted to take several utterances and to focus on emotion cause words of the last utterance.}. Implementation details about Emp-RFT and baselines are covered in Appendix \ref{sec:ergm}.
	
	\begin{table*}[hbt!]
		\centering
		\resizebox{0.98\textwidth}{!}{%
			{\tiny
				\begin{tabular}{ccccc|cccc}
					\specialrule{.08em}{.05em}{.2em}
					\multirow{2}{*}{\textbf{Method}}                 & \multicolumn{4}{c}{\textbf{Automatic Evaluation}}       &   \multicolumn{4}{c}{\textbf{Human Evaluation}}       \\ 
					& \textbf{PPL} & \textbf{Dist-1} & \textbf{Dist-2} & ${\textbf{F}}_{\textbf{BERT}}$ & \textbf{Empathy} & \textbf{Relevance} & \textbf{Fluency} & \textbf{Diversity}\\ 
					\specialrule{.05em}{.05em}{.2em}
					\multicolumn{1}{l}{\textbf{MoEL}}        & 38.04                  & 0.44        & 2.10                          & 0.11        & 3.25       & 3.73                 & 3.49    & 2.85              \\
					\multicolumn{1}{l}{\textbf{EmpDG}}       & 37.29                  & 0.46       & 2.02                        &   0.14      & 3.30       & 3.76                 & 3.57     & 3.11             \\
					\multicolumn{1}{l}{\textbf{MIME}}       & 37.09                 & 0.47       & 1.91                          &  0.13    & 3.23       & 3.78                 & 3.53     & 2.83              \\
					\multicolumn{1}{l}{\textbf{MIME+Focused S1}}       & 36.43                  & 0.52        & 2.21                  & 0.15       & 3.34                & 3.84      & 3.65                 & 3.15                   \\
					\multicolumn{1}{l}{\textbf{Blender+Focused S1}}        & \textbf{13.21}$^{*}$                 & 3.11$^{*}$        & 4.38$^{*}$     &    0.31$^{*}$     & 3.69$^{*}$      & 4.11$^{*}$               & 4.05$^{*}$     & 3.78$^{*}$              \\
					\specialrule{.08em}{.05em}{.2em}
					\multicolumn{1}{l}{\textbf{Emp-RFT}}           & 13.59$^{*}$       & \textbf{3.24$^{*}$}                 & \textbf{4.59}$^{*}$        & \textbf{0.34}$^{*}$         & \textbf{3.78}$^{*}$                & \textbf{4.23}$^{*}$      & \textbf{4.11}$^{*}$              & \textbf{4.02}$^{*}$   \\
					\multicolumn{1}{l}{\textbf{w/o FTR}}          & 15.22        & 3.22$^{*}$                 & 4.49$^{*}$        & 0.27                 & 3.56                & 4.05     & 4.01          & 3.95$^{*}$   \\
					\multicolumn{1}{l}{\textbf{w/o CP}}           & 13.89$^{*}$        & 3.07               & 4.36        & 0.33$^{*}$                & 3.74$^{*}$                & 4.20$^{*}$       & 4.05$^{*}$              & 3.84$^{*}$    \\
					\multicolumn{1}{l}{\textbf{w/o (CP+NEKD)}}           &  14.87      & 2.89               & 4.08      & 0.28               & 3.61                & 4.02       & 3.95            & 3.69    \\
					\multicolumn{1}{l}{\textbf{w/o (CP+NEKD+FCK)}}           & 15.45        & 2.75                & 3.86       & 0.23         & 3.51                & 3.89       & 3.83              & 3.50    \\
					\specialrule{.08em}{.05em}{.2em}
					\specialrule{.08em}{.05em}{.2em}					
					\multicolumn{1}{l}{\textbf{MoEL}}        & 41.13                  & 0.40        & 1.96                          & 0.08        & 2.97       & 3.44                 &3.30    & 2.55              \\
					\multicolumn{1}{l}{\textbf{EmpDG}}       & 40.10                  & 0.41        & 1.91                        &   0.11      & 3.01       & 3.50                 & 3.32     & 2.85              \\
					\multicolumn{1}{l}{\textbf{MIME}}       & 40.51                  & 0.42        & 1.82                           &  0.09      & 2.94       & 3.51                 &3.29     & 2.53              \\
					\multicolumn{1}{l}{\textbf{MIME+Focused S1}}       & 39.58                  & 0.48        & 2.11                  & 0.11       &3.05                 & 3.59       & 3.39                 & 2.91                   \\
					\multicolumn{1}{l}{\textbf{Blender+Focused S1}}        & 16.96                 & 3.03$^{*}$        & 4.19$^{*}$     &    0.26$^{*}$      & 3.43$^{*}$       & 3.90$^{*}$                 & 3.88$^{*}$     &3.65$^{*}$              \\
					\specialrule{.08em}{.05em}{.2em}
					\multicolumn{1}{l}{\textbf{Emp-RFT}}           & \textbf{14.71}$^{*}$        & \textbf{3.21}$^{*}$                 & \textbf{4.48}$^{*}$        & \textbf{0.32}$^{*}$         & \textbf{3.66}$^{*}$                & \textbf{4.15}$^{*}$       & \textbf{4.01}$^{*}$              & \textbf{3.91}$^{*}$    \\
					\multicolumn{1}{l}{\textbf{w/o FTR}}          & 17.12        & 3.20$^{*}$                 & 4.40$^{*}$        & 0.22                    & 3.32                & 3.83       & 3.85              &3.88$^{*}$    \\
					\multicolumn{1}{l}{\textbf{w/o CP}}           &15.12$^{*}$        & 3.04$^{*}$                 & 4.28$^{*}$        & 0.31$^{*}$                & 3.62$^{*}$                & 4.11$^{*}$       & 3.96$^{*}$              & 3.71$^{*}$    \\
					\multicolumn{1}{l}{\textbf{w/o (CP+NEKD)}}           &  16.24       & 2.84                 & 4.02        & 0.25$^{*}$               & 3.50$^{*}$                & 3.92$^{*}$       & 3.84              & 3.61    \\
					\multicolumn{1}{l}{\textbf{w/o (CP+NEKD+FCK)}}           & 17.33        & 2.71                 & 3.78        & 0.20           & 3.42                & 3.82       & 3.76              & 3.42    \\
					\specialrule{.08em}{.05em}{.2em}
					
				\end{tabular}%
			}
		}
		\caption{Results of automatic evaluation and human ratings on all(top) and multi-turn(bottom) instances. * means superior results with $p$-value < 0.05 (sign test).}
		\label{table:eval1}
	\end{table*}

	\begin{table}[hbt!]
		\centering
		\resizebox{\columnwidth}{!}{%
			{\small
				\begin{tabular}{cccc}
					\specialrule{.08em}{.05em}{.2em}
					\textbf{Emp-RFT vs.} & \textbf{Win (\%)} & \textbf{Lose (\%)} & $\boldsymbol{\kappa}$  \\ 
					\specialrule{.08em}{.05em}{.2em}
					\multicolumn{1}{l}{\textbf{MoEL}}    & 74.4/82.2            & 9.3/6.9                  & 0.67/0.73        \\
					\specialrule{.08em}{.05em}{.2em}
					\multicolumn{1}{l}{\textbf{EmpDG}}    & 70.3/77.7           & 12.3/9.8                  & 0.61/0.70        \\  
					\specialrule{.08em}{.05em}{.2em}  
					\multicolumn{1}{l}{\textbf{MIME}}    & 71.6/79.5            & 11.1/8.2                  & 0.64/0.71        \\
					\specialrule{.08em}{.05em}{.2em}  
					\multicolumn{1}{l}{\textbf{MIME+Focused S1}}    & 65.3/74.5            & 13.2/10.8       & 0.61/0.66        \\ 
					\specialrule{.08em}{.05em}{.2em}  
					\multicolumn{1}{l}{\textbf{Blender+Focused S1}}    & 32.0/38.6           & 25.5/22.5                  & 0.46/0.48        \\
					\specialrule{.08em}{.05em}{.2em}
				\end{tabular}%
			}
		}
		\caption{Results of human A/B test. The results in front of and behind ‘/' are each on all instances and multi-turn instances. Fleiss' kappa ($\boldsymbol{\kappa}$) denotes agreements among human workers, where $0.4 < \boldsymbol{\kappa} < 0.6$ and $0.6 < \boldsymbol{\kappa} < 0.8$ indicate moderate and substantial agreements, respectively.}
		\label{table:eval2}
	\end{table}	
	
	\subsection{Evaluation Metrics}
	\textbf{Automatic Evaluation.} We evaluated the models, using the following three metrics: (1) Perplexity (\textbf{PPL}) \cite{vinyals2015neural} measures how highly likely tokens are generated, which evaluates the overall quality of the model. (2) Distinct-n (\textbf{Dist-n}) \cite{li2016diversity} measures how diverse the generated response is via the unique words within its n-gram. (3). We use BERTscore ($\textbf{F}_{\textbf{BERT}}$) \cite{zhang2019bertscore} which measures token-level semantic similarities between the generated response and the gold response based on embeddings from BERT \cite{devlin2019bert}.
	
	\textbf{Human Ratings.} Human evaluations for the dialogues models are essential because of insufficient reliability on automatic metrics. We randomly sampled 100 test dialogues and asked 3 human workers to score models' generated responses on 1 to 5 point scale, following the four metrics \cite{rashkin2019towards}: (1) \textbf{Empathy} measures whether the generated response understands the speaker's emotion and situation. (2) \textbf{Relevance} measures whether the generated response is coherent to the context. (3) \textbf{Fluency} measures whether the generated response is grammatically correct and readable. (4) Since we conclude that models generating generic responses are not empathizing to the speaker, we use \textbf{Diversity} to measure whether the generated response is non-generic.

	\textbf{Human A/B Test.} We further conducted a human A/B test which provides stronger intuitions and higher agreements than human ratings, because this is carried with 3 human workers selecting the better response when given two generated responses \cite{sabour2021cem}.
	
	\subsection{Analysis of Response Generation}
	We abbreviate feature transition recognizer, contrastive PPLM, next emotion and keywords detection, and fusing context with keywords as FTR, CP, NEKD, and FCK, respectively.
	
	\textbf{Automatic Evaluation Results.} The overall automatic evaluation results are shown in the left part of Table \ref{table:eval1}. Emp-RFT performed exceedingly on all metrics except for PPL, which was nearly the same as Blender+Focused S1. The improvements on other metrics indicated that our approach was effective for generating generally high quality and non-generic responses which were also semantically similar with the gold response. While the utilization of pretrained models yielded significant improvements compared to models only trained on EmpatheticDialogues, Emp-RFT showed even greater performance when compared to Blender+Focused S1 endowed with more significant number of dialogues. In addition, due to utilization of FTR, Emp-RFT obtained remarkable results even on multi-turn instances, whereas, other models suffered due to their means of utilizing features for the context at the coarse-grained level.
	
	\textbf{Human Evaluation Results.} In the right part of Table \ref{table:eval1}, Emp-RFT acquired the highest scores on all metrics, which demonstrated that all components of Emp-RFT helped generate responses that are empathetic, coherent to the context, and non-generic. Also, utilizing pretrained models showed significant improvements, especially on Fluency and Diversity scores. In Table \ref{table:eval2}, the generated responses from Emp-RFT were more preferred, which indicated Emp-RFT consistently outperformed other methods in various experiments. When observing at the models' performance difference between multi-turn instances and all instances, only Emp-RFT continued to perform consistently, whereas other models showed significant performance drops under multi-turn instances. From this, we concluded that Emp-RFT continuously understood the dialogue flow.

	\textbf{Ablation Study.} To better understand effects of each component in Emp-RFT, we conducted the ablation study. We gradually ablated each component within the response generation strategy in a hierarchical manner. (1) \textbf{w/o FTR}: Feature transition recognizer was disabled, which resulted in considerable drops on all metrics, especially on PPL, $\mathrm{F}_{\mathrm{BERT}}$, Empathy, and Relevance scores on multi-turn instances, because Emp-RFT could not grasp the attention-needed features of utterance within multi-turn instances through FTR. (2) \textbf{w/o CP}: Contrastive PPLM was removed, which caused lower Dist-n and Diversity scores, because Emp-RFT could not actively use various $\hat{k}^{y}$ when generating responses through CP. (3) \textbf{w/o (CP+NEKD)}: Next emotion and keywords detection were disabled, which interfered with Emp-RFT's utilization of the next emotion and keyword. It droped not only Dist-n and Diversity scores but also other metrics.  (4) \textbf{w/o (CP+NEKD+FCK)}: Fusing the context representation with keyword representations was disregarded. Since keywords were no longer emphasized for context modelling, such information could not get more attention when generating responses. It caused drops on all metrics, particularly on $\mathrm{F}_\mathrm{BERT}$, Dist-n, and Diversity.
	
	\begin{table}[t]
		\centering
		\resizebox{\columnwidth}{!}{%
			{\small
				\begin{tabular}{ccccc}
					\specialrule{.08em}{.05em}{.2em}
					\textbf{Method} & \textbf{Top-1 Acc} & \textbf{Top-5 Acc} & \textbf{macro-F1}  \\ 
					\specialrule{.08em}{.05em}{.2em}
					\multicolumn{1}{l}{\textbf{CoMAE}}         & 41.12            & 80.09            & 39.61   \\
					\multicolumn{1}{l}{\textbf{Emp-RFT}}           & \textbf{42.08}            & \textbf{80.78}            & \textbf{40.39}  \\
					\specialrule{.08em}{.05em}{.2em}
					\specialrule{.08em}{.05em}{.2em}
					\textbf{Method} & \textbf{TL-P} & \textbf{TL-R} & \textbf{TL-F1}  \\ 
					\specialrule{.08em}{.05em}{.2em}
					\multicolumn{1}{l}{\textbf{ConceptFlow}}        & 37.68           & 48.27           & 42.32         \\
					\multicolumn{1}{l}{\textbf{CG-nAR}}           & 44.62           & 61.94           & 51.87      \\
					\multicolumn{1}{l}{\textbf{Emp-RFT}}           & \textbf{45.35}           & \textbf{63.15}           & \textbf{52.78}         \\
					\specialrule{.08em}{.05em}{.2em}
				\end{tabular}%
			}
		}
		\caption{The results of next emotion(top) and keywords(bottom) detection. We used the metrics introduced in Table \ref{table:anno1}.}
		\label{table:nekdresults}
	\end{table}

	\subsection{Analysis of Next Emotion and Keywords} 
	We report the results in terms of NEKD in Table \ref{table:nekdresults}. Since all baselines have not conducted NEKD,  we trained models showing promising results such as \textbf{CoMAE} \cite{zheng-etal-2021-comae}, \textbf{ConceptFlow} \cite{zhang2020grounded} and \textbf{CG-nAR} \cite{zou2021thinking} with EmpatheticDialogues. (More details are covered in Appendix \ref{sec:ned}). Then, we compared Emp-RFT to those models. Emp-RFT outperformed other models on all metrics, which proved Emp-RFT figured out which emotion and keywords were proper for generation.
	
	\begin{table}[hbt!]
		\centering
		\resizebox{\columnwidth}{!}{%
			{\small
				\begin{tabular}{c}
					
					\specialrule{.08em}{.05em}{.2em}
					\multicolumn{1}{l}{\textbf{Emotion Label}: Furious} \\ 
					\multicolumn{1}{l}{\textbf{Annotated Emotion}: Annoyed $\rightarrow$ Apprehensive $\rightarrow$ Confident}\\
					\multicolumn{1}{l}{$\rightarrow$ Hopeful}\\
					\multicolumn{1}{l}{$u^{1}$: My \textcolor{red}{roommate} \textcolor{red}{eats} my \textcolor{red}{food} \textcolor{red}{sometimes}. This \textcolor{red}{makes} me so \textcolor{red}{angry}!} \\ 
					\multicolumn{1}{l}{$u^{2}$: You should get a \textcolor{blue}{mini} \textcolor{blue}{fridge} and put it in your \textcolor{blue}{room}, with a \textcolor{blue}{lock} on it.} \\  
					\multicolumn{1}{l}{$u^{3}$: I \textcolor{red}{think} that's a \textcolor{red}{great}  \textcolor{red}{idea}. I \textcolor{red}{know} where to get those \textcolor{red}{fridges} for \textcolor{red}{cheap}.} \\
					\multicolumn{1}{l}{\textbf{Gold}: Yea man go for it, don't \textcolor{blue}{procrastinate}.} \\
					\specialrule{.08em}{.05em}{.2em}
					\multicolumn{1}{l}{\textbf{MoEL}: I am sorry you have to hear that. I hope it works out for you.} \\
					\multicolumn{1}{l}{\textbf{EmpDG}: I agree with you. I think it isn't worth before you get it back.} \\
					\multicolumn{1}{l}{\textbf{MIME}: I am sorry to hear that. I hope you don't have to deal with that.} \\
					\multicolumn{1}{l}{\textbf{MIME+Focused S1}: I agree. I have a friend who is not to be a parent.} \\
					\multicolumn{1}{l}{\textbf{Blender+Focused S1}: Roommates can be so annoying.} \\
					\specialrule{.08em}{.05em}{.2em}
					\multicolumn{1}{l}{$\hat{e}^{y}$: Trusting, $\hat{k}^{y}$: procrastinate, safety, profit} \\
					\multicolumn{1}{l}{\textbf{Emp-RFT}: Don't procrastinate. It makes your foods safety.} \\
					\multicolumn{1}{l}{\textbf{w/o FTR}: I agree. I have a safety but pricy fridge.} \\
					\specialrule{.08em}{.05em}{.2em}
					
					\specialrule{.08em}{.05em}{.2em}
					\multicolumn{1}{l}{\textbf{Emotion Label}: Proud} \\ 
					\multicolumn{1}{l}{\textbf{Annotated Emotion}: Trusting $\rightarrow$ Proud $\rightarrow$ Grateful$\rightarrow$ Sentimental} \\ 
					\multicolumn{1}{l}{$\rightarrow$ Disappointed $\rightarrow$ Sentimental} \\
					\multicolumn{1}{l}{$u^{1}$: Hi, My \textcolor{red}{girlfriend} was recently \textcolor{red}{let} go of her job, but she put so much} \\ 
					\multicolumn{1}{l}{work in to finding a \textcolor{red}{new} \textcolor{red}{job} and \textcolor{red}{got} one with better \textcolor{red}{pay}!} \\ 
					\multicolumn{1}{l}{$u^{2}$: That's great! It was \textcolor{blue}{meant} to happen then.} \\  
					\multicolumn{1}{l}{$u^{3}$: \textcolor{red}{Yes} I \textcolor{red}{agree}, she was so \textcolor{red}{happy} when she \textcolor{red}{found} out the \textcolor{red}{job} was hers.} \\ 
					\multicolumn{1}{l}{$u^{4}$: I would have been too. I've never been \textcolor{blue}{fired}. I always \textcolor{blue}{quit}.} \\
					\multicolumn{1}{l}{$u^{5}$: She wasn't \textcolor{red}{really} fired, her position was no more. \textcolor{red}{Schools} don't} \\
					\multicolumn{1}{l}{have \textcolor{red}{much} room \textcolor{red}{left} for Home \textcolor{red}{Ec} teachers. \textcolor{red}{Sad}.} \\
					\multicolumn{1}{l}{\textbf{Gold}: I \textcolor{blue}{remember} taking Home \textcolor{blue}{Ec}. I loved that \textcolor{blue}{class}.} \\
					\specialrule{.08em}{.05em}{.2em}
					\multicolumn{1}{l}{\textbf{MoEL}: I am sorry to hear that. I hope you can find a new job.} \\
					\multicolumn{1}{l}{\textbf{EmpDG}: I am sorry you had that. I am not a fan of the past few years} \\
					\multicolumn{1}{l}{ago. I am not sure she had a close relationship.} \\
					\multicolumn{1}{l}{\textbf{MIME}: I am sorry to hear that. I hope you find a new job.} \\
					\multicolumn{1}{l}{\textbf{MIME+Focused S1}: I am sorry to hear that. I hope you find a new job.} \\
					\multicolumn{1}{l}{\textbf{Blender+Focused S1}: She was paid for her hard work.} \\
					\specialrule{.08em}{.05em}{.2em}
					\multicolumn{1}{l}{$\hat{e}^{y}$: Sentimental, $\hat{k}^{y}$: Ec, class, jobs, solution, remember} \\
					\multicolumn{1}{l}{\textbf{Emp-RFT}: I remember the Ec class. It's a relief for her to find the job.} \\
					\multicolumn{1}{l}{\textbf{w/o FTR}: I'm glad she found the solution to the job problem.} \\
					\specialrule{.08em}{.05em}{.2em}
					
					\specialrule{.08em}{.05em}{.2em}
					\multicolumn{1}{l}{\textbf{Emotion Label}: Caring} \\ 
					\multicolumn{1}{l}{\textbf{Annotated Emotion}: Annoyed $\rightarrow$ Caring $\rightarrow$ Sad$\rightarrow$ Caring} \\ 
					\multicolumn{1}{l}{$u^{1}$: My \textcolor{red}{poor} \textcolor{red}{neighbor} is \textcolor{red}{suffering} a \textcolor{red}{lot} \textcolor{red}{without} her \textcolor{red}{husband}.} \\ 
					\multicolumn{1}{l}{$u^{2}$: I \textcolor{blue}{suffer} a lot too when my \textcolor{blue}{wife} is \textcolor{blue}{gone}. What happened to him?} \\  
					\multicolumn{1}{l}{$u^{3}$: He \textcolor{red}{passed} \textcolor{red}{away} from \textcolor{red}{cancer}.} \\ 
					\multicolumn{1}{l}{\textbf{Gold}: Ah, the \textcolor{blue}{evil} \textcolor{blue}{cancer}. Took my \textcolor{blue}{grandmother} as well. I am sure} \\
					\multicolumn{1}{l}{he's off in a better place now.} \\
					\specialrule{.08em}{.05em}{.2em}
					\multicolumn{1}{l}{$\hat{e}^{y}$: Sad, $\hat{k}^{y}$: condolences, cancer, grandmother, evil, lost} \\
					\multicolumn{1}{l}{\textbf{Emp-RFT}: My condolences. I lost my grandmother because of the} \\
					\multicolumn{1}{l}{cancer.} \\
					\multicolumn{1}{l}{\textbf{w/o CP}: I'm sorry to hear that. It's so hard to lost someone.} \\
					\multicolumn{1}{l}{\textbf{w/o (CP+NEKD)}: Oh no. I'm scary to get the cancer.} \\
					\multicolumn{1}{l}{\textbf{w/o (CP+NEKD+FCK)}: Oh no. I'm sorry to hear that.} \\
					\specialrule{.08em}{.05em}{.2em}
					
				\end{tabular}%
			}
		}
		\caption{Model generations. The words marked in red and blue are keywords of the speaker utterance and listener utterance, respectively.}
		\label{table:casestudy}
	\end{table}

	\subsection{Case Study}
	The cases from the models are shown in Table \ref{table:casestudy}. In the first case, MoEL and MIME expressed regret, which was emotionally inappropriate to the context. All baselines except for MoEL failed to grasp the proper features within the context, and therefore generated incoherent responses. Especially, Blender+Focused S1 ignored the features of $u^{3}$. Since Emp-RFT understood the dialogue flow, it became attentive to not only the features of $u^{3}$ but also those of $u^{1}$, $u^{2}$, mentioning (‘procrastinate', ‘foods', ‘safety'), which led to empathy and coherence. In the second case, all baselines couldn't understand the longer context, which resulted in improper empathy. Also, Blender+Focused S1 disregarded the features of $u^{5}$, and therefore overlooked the speaker's sadness. Emp-RFT fully comprehended why the speaker's happiness changed to the sadness. In both cases, without FTR, the responses of Emp-RFT were non-empathetic and incoherent because of dismissing appropriate features. In the third case, we report the case in terms of the response generation strategy. Without CP, NEKD, and FCK, Emp-RFT produced a generic response. With the utilization of FCK, Emp-RFT perceived the word ‘cancer' in $u^{3}$ but expressed excessive emotion by mentioning ‘scary'. When Emp-RFT additionally conducted NEKD, Emp-RFT generated emotionally appropriate responses by mentioning ‘sorry' and ‘hard', and utilized the keyword ‘lost'. Lastly, with CP, Emp-RFT generated a diverse response, actively using $\hat{k}^{y}$.

	\section{Conclusion}
	We proposed a novel approach that recognizes feature transitions between utterances, which led to understanding the dialogue flow and grasping the features of utterance that needs attention. Also, to make our model focus on emotion and keywords related to appropriate features, we introduced a response generation strategy including fusing context with keywords, next emotion and keywords detection, and contrastive PPLM. Experimental results showed that our model outperformed baselines, and especially, achieved significant improvements on multi-turn instances, which proved our approach was effective for empathetic, coherent, and non-generic response generation.

	\section{Ethical Considerations}
	We expect that our proposed approach does not suffer from ethical problems. The dataset we use in our work is EmpatheticDialogues which is English-based. The dataset is constructed by crowdsourcing with Amazon Mechanical Turk, which protects private user information \cite{rashkin2019towards}. In addition, the dialogue dataset is anticipated not to have responses which include discrimination, abuse, bias, etc, because the robust collection procedure of EmpatheticDialogues ensures the quality of the dataset. Thus, we expect that models trained using the dataset, do not generate inappropriate responses which harm the users. However, we inform that our model utilizes a pretrained language model, which may produce inappropriate responses. Lastly, we anticipate our model make potential users be interested and consoled by generating empathetic responses.

	\section*{Acknowledgements}
	
	We thank all anonymous reviewers for their meaningful comments, and Hyeongjun Yang, Chanhee Lee and Sunwoo Kang of Yonsei University for their discussion and feedback about our work. This work was supported by the National Research Foundation of Korea(NRF) grant funded by the Korea government(MSIP; Ministry of Science, ICT \& Future Planning) (No. NRF-2022R1A2B5B01001835). Also, this work was partly supported by the Institute of Information and Communications Technology Planning and Evaluation(IITP) grant funded by the Korean government(MSIT) (No. 2020-0-01361-003, Artificial Intelligence Graduate School Program (Yonsei University)). Kyong-Ho Lee is the corresponding author.

	\bibliography{anthology,custom}
	
	\appendix
	
	\section{Implementation Details}
	\label{sec:Implementationdetail}
	\subsection{Empathetic Response Generation Models}
	\label{sec:ergm}
	We use the official codes of all baselines, and follow the implementations (MoEL \footnote{\url{https://github.com/HLTCHKUST/MoEL}}, EmpDG \footnote{\url{https://github.com/qtli/EmpDG}}, MIME \footnote{\url{https://github.com/declare-lab/MIME}}, MIME+Focused S1 and Blender Focused S1 \footnote{\url{https://github.com/skywalker023/focused-empathy}}). Our model is implemented by Pytorch \footnote{\url{https://pytorch.org/}}, and based on two encoders of BART-base and a decoder of BART-base \footnote{\url{https://huggingface.co/docs/transformers/model_doc/bart}}. Hidden size $d$ is 768 and the number of emotion classes $n_{emo}$ is 32. $MH$ and the number of layers of graph attention network are each 4. Using Adam optimization \cite{DBLP:journals/corr/KingmaB14}, our model is trained on single RTX 3090 GPU with a batch size of 4. We apply early-stopping and select a model showing the best performance through perplexity on the valid set. For contrastive PPLM, we utilize the official code of PPLM \footnote{\url{https://github.com/uber-research/PPLM}}. We set a temperature parameter $\tau$ and batch size to 0.5 and 64, respectively. Through represenations derived from the last token of BART decoder whose parameters are frozen, we can obtain each response representation $r_{a}$ and each keyword set representation $ks_{a}$, where the keyword set corresponds to the response. Thus, $ks_{a}$ becomes a positive sample for $r_{a}$, and keyword set representations for other responses in the same batch become negative samples.
	\subsection{Next Emotion and Keywords Detection}
	We utilize the repositories and follow implemetation details of CoMAE \footnote{\url{https://github.com/chujiezheng/CoMAE}}, ConceptFlow \footnote{\url{https://github.com/thunlp/ConceptFlow}}, and CG-nAR \footnote{\url{https://github.com/RowitZou/CG-nAR}}. We train three models, using EmpatheticDialogues instead of originally used datasets.
	\label{sec:ned}

\end{document}